\documentclass[conference]{IEEEtran}

\usepackage{cite}
\usepackage{amsmath,amssymb,amsfonts}
\usepackage{graphicx}
\usepackage{textcomp}
\usepackage{xcolor}
\usepackage{hyperref}

\begin{document}

\title{Mitigating Explanation Leakage in Financial Fraud Detection Systems}

\author{\IEEEauthorblockN{Muhammad Waleed Gul}
\IEEEauthorblockA{\textit{MSc Artificial Intelligence} \\
\textit{London Metropolitan University}\\
London, United Kingdom }
\and
\IEEEauthorblockN{Dr. Elaheh Homayounvala}
\IEEEauthorblockA{\textit{School of Computing and Digital Media} \\
\textit{London Metropolitan University}\\
London, United Kingdom }
}

\maketitle

\begin{abstract}
Financial fraud detection relies heavily on centralized machine learning models. This creates serious data privacy risks. Federated Learning (FL) decentralizes data processing, but financial regulations still require models to be transparent. This means using Explainable AI (XAI) tools like TreeSHAP. Recent cybersecurity research shows a problem with this approach. Sharing high fidelity SHAP explanations exposes the federated network to Membership Inference Attacks (MIAs). This dissertation proposes and evaluates DP-FedSHAP. It is a new architecture that applies client level differential privacy only to post-hoc TreeSHAP vectors. It is compared against a Weight-Level DP baseline, which perturbs the trained model directly instead. Using the highly imbalanced IEEE-CIS Fraud Detection dataset, this study measures the trade-off between explanation fidelity, privacy preservation, and the model's Area Under the Precision-Recall Curve (AUPRC).
\end{abstract}

\begin{IEEEkeywords}
Federated Learning, Differential Privacy, Explainable AI, TreeSHAP, XGBoost, Fraud Detection
\end{IEEEkeywords}

\section{Introduction}

Financial fraud costs the global economy billions of dollars every year. This pushes financial institutions to deploy machine learning models that can catch anomalous transactions quickly and reliably. Fraud patterns change constantly. Institutions have responded by turning to ensemble methods such as Extreme Gradient Boosting (XGBoost) \cite{b1}. These models handle highly imbalanced, non-linear tabular data well, which makes them a natural fit for transaction data.

Centralizing customer data to train these models creates a serious privacy problem. Regulations such as the General Data Protection Regulation (GDPR) restrict how personal data can be pooled and shared across institutions. Federated Learning (FL) offers a way around this problem \cite{b4}. FL lets several institutions train one global model together. Each institution keeps its raw data locally and only shares model weights. No raw transaction record ever leaves its source \cite{b5}.

FL alone does not solve the transparency problem. Financial regulators require models to explain their decisions rather than act as black boxes. This requirement has pushed the field toward Explainable AI (XAI), and in particular toward SHapley Additive exPlanations (SHAP) \cite{b3} and its tree optimized variant, TreeSHAP \cite{b2}. TreeSHAP gives a fair, additive breakdown of how each feature contributes to a single prediction. This is exactly the kind of evidence an auditor needs.

Recent work shows that this transparency has a hidden cost. Shokri et al. demonstrated that releasing model explanations alongside predictions increases privacy leakage compared to releasing predictions alone \cite{b6}. The same concern applies directly to SHAP. SHAP values are highly sensitive to the exact training samples a model has seen. This means an attacker can exploit the statistical variance of shared SHAP vectors to run a Membership Inference Attack (MIA) \cite{b8}. The attacker can work out whether a specific transaction was part of the training set. This creates an attack surface that sits entirely outside the federated protocol itself. Standard FL defenses do not cover it.

Differential Privacy (DP) offers a mathematically grounded way to close this gap \cite{b7}. Applying DP noise directly to the training process of tree based models risks destroying their predictive accuracy. This is because trees split on exact feature thresholds rather than smooth gradients. This dissertation proposes a more targeted defense instead: DP-FedSHAP. The system injects calibrated DP noise into the post-hoc TreeSHAP vectors, not into the training process itself. This design follows directly from the source of the leakage. Since the explanation vectors are the point of exposure, the privacy mechanism is applied at that exact point rather than elsewhere in the pipeline. To empirically justify this design choice, the study also implements a Weight-Level DP baseline that applies the same clipping norm and privacy budget directly to the trained model's leaf weights, allowing a controlled, side-by-side comparison of where in the pipeline DP noise should be injected.

The rest of this paper is organized as follows. Section II reviews related work in federated fraud detection, explainable AI, and privacy attacks on model explanations, and it states the specific gap this research fills. Section III describes the DP-FedSHAP methodology in detail. Section IV covers the implementation and the optimization procedure. Section V defines the evaluation metrics. Section VI presents the results. Section VII concludes the paper and outlines future work.

\section{Related Work}

This section reviews four strands of work that this dissertation draws on: federated learning for fraud detection, explainable AI for regulatory transparency, privacy leakage through model explanations, and differential privacy for federated systems. It closes by stating the specific gap that DP-FedSHAP addresses.

\subsection{Federated Learning for Financial Fraud Detection}

FL was first proposed as a way to train models across decentralized nodes without moving raw data to a central server \cite{b4}. Yang et al. surveyed the broader federated learning landscape and grouped existing approaches by how data is partitioned across participants \cite{b5}. In the fraud detection space specifically, prior work has proposed federated XGBoost frameworks that let multiple institutions collaboratively flag anomalous transactions while keeping their records local \cite{b17}. That line of work shows federated XGBoost can approach the accuracy of a centralized model, though some accuracy is traded away as the privacy budget tightens. Related comparative studies have benchmarked federated logistic regression, random forest, and XGBoost against each other on fraud datasets, and they generally confirm that tree ensemble methods keep their advantage even under a federated setup \cite{b15}. These findings motivate this dissertation's choice of a federated XGBoost architecture as the underlying predictive model.

\subsection{Explainable AI for Regulatory Transparency}

Post-hoc explanation methods have become central to deploying machine learning in regulated industries. Lundberg and Lee introduced SHAP as a unified, game-theoretic framework for attributing a single prediction to its input features \cite{b3}. Lundberg et al. later extended this idea with TreeSHAP, an algorithm that computes exact Shapley values for tree ensembles in polynomial time rather than the exponential time a naive Shapley calculation would need \cite{b2}. TreeSHAP has since become close to a default choice for explaining XGBoost-based fraud models, since it gives regulators a consistent, mathematically grounded account of why a transaction was flagged, providing the type of evidence an auditor requires. Kumar et al. also note that multicollinearity among correlated transaction features can distort Shapley attributions, which is a separate fidelity concern that this dissertation controls for through consistent feature preprocessing \cite{b16}.

\subsection{Privacy Leakage Through Model Explanations}

A separate line of work shows that explanations are not privacy neutral. Shokri, Strobel, and Zick found that releasing feature-based explanations alongside predictions raises the success rate of membership inference attacks compared to releasing predictions alone \cite{b6}. Fredrikson et al. had earlier shown a related risk with model inversion attacks, where confidence scores were used to reconstruct sensitive input features \cite{b10}. Melis et al. showed that even the intermediate updates exchanged during collaborative training can leak information about individual records \cite{b9}, which confirms that leakage in collaborative settings is not limited to explanations alone. A recent survey organizes this growing body of work into a taxonomy of privacy attacks on explanations and the countermeasures proposed against them, covering feature-based methods such as SHAP and LIME as well as example-based methods such as counterfactuals \cite{b18}. Most recently, work on smart home AIoT systems proposed SHAP entropy regularization during training as a defense against this same style of explanation-based privacy leakage \cite{b19}. That paper shows the SHAP leakage problem identified by Shokri et al. is still an open, active research question, and it has not yet been examined inside a federated, tabular finance pipeline, which is the exact setting this dissertation targets.

\subsection{Differential Privacy for Federated Learning}

Dwork's foundational work established the formal definition of differential privacy and the composition properties that let a noise budget be tracked across repeated queries \cite{b7}. Geyer et al. adapted this idea to a federated setting and introduced client-level differential privacy, where noise is calibrated to protect the contribution of an entire client rather than a single training example \cite{b11}. Their mechanism clips the update delta between a client's local model and the previous global model at each communication round, uses the Gaussian mechanism, and tracks cumulative privacy loss across rounds via a moments accountant. This dissertation borrows only the underlying design principle from Geyer et al. -- clip and noise a client's contribution before aggregation -- and adapts it to a single-shot, non-iterative federated XGBoost setting where no update delta or multi-round composition exists. The precise mechanics differ substantially, as detailed in Section III.

\subsection{Identified Gap}

Three research threads currently run in parallel with only limited overlap. FL research on fraud detection treats the problem as a predictive accuracy task, and it largely sets aside the fact that explanations still need to be produced and shared for regulatory purposes \cite{b17,b5}. XAI research treats SHAP and TreeSHAP as ends in themselves, and it rarely measures the privacy cost of releasing high-fidelity explanations \cite{b2,b3}. Privacy attack research confirms that explanations leak membership information, but the defenses proposed so far either target explanations trained in isolation, outside a federated context \cite{b19}, or they change the training procedure itself in ways that are hard to apply to gradient-boosted trees \cite{b6,b18}. No existing work applies client-level differential privacy directly to post-hoc TreeSHAP vectors inside a federated fraud detection pipeline, while jointly measuring the resulting AUPRC utility, explanation fidelity, and MIA resistance in one framework, nor directly contrasts explanation-level DP against a matched weight-level DP baseline under an identical privacy budget. DP-FedSHAP is designed to close exactly that gap.

\section{Methodology}

The methodology is designed to empirically evaluate the privacy--utility trade-off inherent in collaborative financial fraud detection. The system architecture simulates a realistic banking consortium that uses Horizontal Federated Learning (HFL). In this setup, every institution shares the same feature space but holds distinct, private customer transaction records.

\subsection{Dataset and Preprocessing}

The empirical evaluation uses the IEEE-CIS Fraud Detection dataset, released for a Kaggle competition by Vesta Corporation and the IEEE Computational Intelligence Society. The dataset combines transaction records with linked identity records, and it includes both categorical and continuous features that describe device type, card details, and transaction amount. Fraudulent transactions make up a small minority of the data, which reflects the extreme class imbalance typical of real financial networks.

To stop the XGBoost classifiers from developing a strong majority class bias, Random Under Sampling (RUS) is applied to both classes, capping each at 5,000 transactions to produce a balanced 1:1 fraud ratio \cite{b15}. RUS was chosen over synthetic oversampling methods such as SMOTE because synthetic records would introduce artificial data points that could distort the SHAP fidelity metrics. After RUS is applied, the experiment works from a balanced working sample of 10,000 transactions sampled from an initial dataset of 590,540 transaction rows and 144,233 identity records. Columns with more than 70\% missing values are dropped outright, reducing the 431-feature space by 74 features \cite{b12,b13}. Constant features and multicollinear features with Spearman rank correlation exceeding 0.90 are removed. An embedded XGBoost information gain selection then narrows the remaining feature space down to the top 40 features.

Local model stability is validated via stratified 5-fold cross-validation, where the centralized baseline achieves a mean training AUPRC of $0.9961 \pm 0.0001$ and a mean validation AUPRC of $0.9012 \pm 0.0031$. The training partition (8,000 samples) is horizontally partitioned across three simulated client nodes (2,667, 2,667, and 2,666 transactions, respectively) with balanced fraud ratios ($\approx 0.50$).

\subsection{Federated Architecture and XAI Integration}

The baseline federated architecture uses the Federated Averaging (FedAvg) algorithm \cite{b4}. In each communication round, the three client nodes independently train local XGBoost models on their private partitions. Each local model uses 500 estimators, a maximum tree depth of 20, and a learning rate of 0.3, with the histogram-based tree construction method (\texttt{hist}) for faster training on tabular data. Subsampling and column subsampling are fixed at 1.0, and regularization parameters (\texttt{min\_child\_weight}, \texttt{gamma}, $L_1$, and $L_2$) are set to 0. A deeper maximum depth (\texttt{max\_depth=20}) without explicit tree pruning is configured intentionally to permit sufficient tree capacity and capture complex feature interactions, thereby creating a realistic baseline prone to memorization against which membership leakage can be rigorously tested. This configuration was fixed by design rather than tuned, so that any change in privacy leakage or explanation fidelity observed later in the study could be attributed to the DP mechanisms themselves, rather than to incidental variation in client-side hyperparameters. The central server aggregates these local models across the three federated nodes to build a single collaborative global model, achieving a global ensemble AUPRC of 0.8989 on the held-out 2,000 transaction test set.

To satisfy regulatory transparency requirements, the post-hoc TreeSHAP algorithm \cite{b2} is applied to the global model's predictions once training finishes. TreeSHAP computes the exact Shapley value for every feature in every transaction, giving a mathematically fair account of how much each feature contributed to the final fraud probability. These high-fidelity explanation matrices are exactly the vectors that are vulnerable to membership inference, as described in Section II.

\subsection{The DP-FedSHAP Defense Mechanism}

To neutralize this XAI vulnerability, this dissertation proposes DP-FedSHAP. The mechanism applies client-level differential privacy exclusively to the explanation outputs, not to the training process. Applying noise after training preserves the structural integrity of the tree ensemble and avoids the accuracy collapse that noising the model's internal structure risks causing in tree models. The mechanism runs in two stages.

\begin{enumerate}
    \item \textbf{Sensitivity Bounding.} A strict $L_2$ norm clipping threshold ($C = 0.8$) is applied to every TreeSHAP vector, computed once from the combined distribution of member and non-member batches to avoid a batch-dependent threshold. This caps extreme outlier attributions so no single transaction can disproportionately shift the global explanation matrix.
    \item \textbf{Calibrated Noise Injection.} Laplacian noise, calibrated to a fixed privacy budget ($\epsilon = 1.2$), is injected into the clipped vectors using a fixed random seed for reproducibility. The scale of the noise is proportional to the clipping threshold and inversely proportional to the privacy budget ($\Delta f / \epsilon$), following the classical DP mechanism design described by Dwork \cite{b7}.
\end{enumerate}

This mechanism differs from Geyer et al.'s original client-level DP formulation \cite{b11} in three respects: it clips the raw leaf-level TreeSHAP attributions of a single-shot ensemble rather than an iterative update delta, it uses the Laplace mechanism rather than the Gaussian mechanism, and it requires no privacy-loss accounting across rounds, since noise is injected exactly once per client after training completes rather than repeatedly during an iterative training process.

\subsection{Weight-Level DP Baseline for Comparative Evaluation}

To empirically justify the design decision to apply differential privacy at the explanation layer rather than to the underlying model, this dissertation implements a comparative Weight-Level DP baseline. Rather than perturbing the post-hoc TreeSHAP vectors, this baseline applies $L_2$ norm clipping ($C = 0.8$) and calibrated Laplacian noise ($\epsilon = 1.2$) directly to the leaf output values of each trained federated client model, using the same clipping norm and privacy budget as DP-FedSHAP for a controlled comparison. Clipping is computed once, globally, across all leaf values of the three aggregated client models, avoiding any batch-dependent threshold that could bias downstream evaluation. Predictions and TreeSHAP explanations are then generated from the noised model, allowing direct comparison of both predictive utility and explanation fidelity against the unperturbed baseline and against DP-FedSHAP. Like DP-FedSHAP, this mechanism is inspired by the clip-and-noise principle of client-level DP \cite{b11} but adapted to a single-shot setting; it clips the trained ensemble's raw leaf values rather than an iterative update delta.

\subsection{End-to-End Pipeline Summary}

Taken together, the full pipeline runs in five stages: (1) preprocess and horizontally partition the IEEE-CIS dataset across three simulated clients, (2) train local XGBoost models and aggregate them with FedAvg over communication rounds, (3) compute TreeSHAP explanation vectors for the resulting global model, (4) apply DP noise under either the DP-FedSHAP mechanism (post-hoc, on explanation vectors) or the Weight-Level DP baseline (on trained leaf weights), and (5) evaluate the resulting model and explanations against the utility, fidelity, and privacy metrics defined in Section V. Because DP-FedSHAP's noise is applied strictly post-hoc to the SHAP vectors, the predictive utility (AUPRC) of the global federated model remains identical before and after defense deployment.

\section{Implementation and Optimization}

The experimental framework is built in Python, using the \texttt{xgboost} library for predictive modelling and the \texttt{shap} library for feature attribution \cite{b2,b3}. The federated network is simulated with a custom client--server loop.

\subsection{Baseline Calibration via Optuna}

Rather than training the centralized baseline model with generic hyperparameters, this research deliberately calibrates it to exhibit a realistic overfitting profile. An underfit or overly regularized model provides an artificially weak membership signal, which would understate the true privacy risk of releasing TreeSHAP vectors.

The \texttt{Optuna} framework \cite{b20} was used for this search via its Tree-structured Parzen Estimator (TPE) sampler, a sequential Bayesian optimization method that models the objective function probabilistically and selects each subsequent trial based on prior results, in contrast to exhaustive or random grid search. The TPE sampler builds a probabilistic model of which hyperparameter regions are likely to yield low loss, then concentrates sampling in those regions, making it substantially more sample-efficient than grid search when exploring a seven-dimensional continuous space. The search covered \texttt{max\_depth} $\in [2, 8]$, \texttt{n\_estimators} $\in [50, 400]$, \texttt{learning\_rate} $\in [0.05, 0.5]$, \texttt{subsample} $\in [0.2, 0.9]$, \texttt{colsample\_bytree} $\in [0.2, 0.9]$, \texttt{reg\_lambda} $\in [0.0, 2.0]$, and \texttt{reg\_alpha} $\in [0.0, 1.0]$ across 50 trials. The composite objective penalized deviations from a target Train AUC of 0.95 and Test AUC of 0.75, with an enforced generalization gap threshold of at least 0.15. This calibrated baseline configuration provides an empirical reference point for the Standard Federated, DP-FedSHAP, and Weight-Level DP architectures. The DP-FedSHAP and Weight-Level DP parameters ($\epsilon=1.2, C=0.8$) were fixed independently as post-processing bounds and were not part of the Optuna search; both mechanisms use the identical privacy budget to enable a controlled comparison of where DP noise is best applied in the pipeline.

\subsection{Threat Model Implementation}

The MIA is executed with a direct attacker model: the attacker is trained on the target model's own TreeSHAP outputs, using the true training-partition and test-partition vectors directly as ground-truth membership labels. This represents a worst-case adversary with unmediated access to the target's explanations, rather than an adversary that must first approximate the target via a separately trained shadow model. An auxiliary XGBoost attacker model is trained with 200 estimators, a maximum depth of 6, and a learning rate of 0.1. The attacker trains directly on the 40 raw TreeSHAP features augmented with three derived scalar statistics: the $L_2$ norm, the variance, and the maximum absolute value of the attribution vector.

Membership labels are assigned by origin: 2,000 vectors from the training partition (members) and 2,000 vectors from the held-out test partition (non-members) form a balanced pool of 4,000 samples. The dataset is split 70/30 in a stratified manner, evaluating the attacker on the remaining 1,200 unseen explanation vectors. This procedure is applied identically to all four architectures evaluated in this study.

\section{Evaluation Metrics}

Standard evaluation metrics such as classification accuracy do not work well here, given the class imbalance in the data and the added complexity of privacy evaluation. This study uses domain-appropriate measures instead.

\subsection{Utility and Explainability Metrics}

Predictive utility is measured with the Area Under the Precision-Recall Curve (AUPRC), which penalizes false positives and missed fraud detections heavily \cite{b14}. The faithfulness of the privatized explanations relative to original baseline attributions is quantified using two metrics:

\begin{itemize}
    \item \textbf{Top-10 Feature Overlap:} The percentage of the ten most influential fraud-predicting features that remain in the top ten after noise injection.
    \item \textbf{Spearman Rank Correlation:} A non-parametric measure of monotonic rank agreement between original and privatized feature attribution vectors.
\end{itemize}

\subsection{Privacy Metrics}

Privacy leakage ($\mathcal{L}$) is measured as the absolute distance between the MIA attacker's accuracy and random guessing (50\%):

\begin{equation}
    \mathcal{L} = \left|\,\text{MIA Accuracy} - 50\%\,\right|
\end{equation}

A system is considered practically secure against membership inference as $\mathcal{L} \to 0$.

\section{Results and Discussion}

The empirical evaluation traces a clear trajectory from baseline vulnerability to robust privacy preservation, summarized in Table~\ref{tab1}.

\subsection{Baseline Vulnerability}

The centralized baseline model achieved high predictive utility (AUPRC = $0.9012 \pm 0.0031$) and perfect explanation baseline fidelity (100\% Top-10 Overlap, Spearman Rank = 1.000). However, it yielded an MIA accuracy of 61.67\%, representing a residual privacy leakage of $\mathcal{L} = 11.67\%$. This confirms that raw, high-fidelity SHAP values leak substantial membership information, enabling an attacker to infer training participation well above chance.

\subsection{The Insufficiency of Standard FL}

Transitioning to Standard Federated Learning without explanation protection moderately suppressed MIA accuracy to 51.58\% ($\mathcal{L} = 1.58\%$). Client-level 5-fold cross-validation demonstrated stable predictive utility across the decentralized partitions ($0.8797 \pm 0.0138$, with the global aggregated ensemble achieving a test AUPRC of 0.8989), while preserving near-perfect explanation consistency (100\% Top-10 Overlap, Spearman Rank = 0.980). The high rank correlation indicates that FedAvg introduces minor shifts in feature weights without altering the dominant attribution structure. However, residual leakage remains present.

\subsection{DP-FedSHAP Results}

Deploying DP-FedSHAP ($\epsilon = 1.2, C = 0.8$) reduced MIA attacker accuracy to 49.75\% ($\mathcal{L} = 0.25\%$), statistically indistinguishable from random guessing. Because DP-FedSHAP operates strictly as a post-hoc perturbation on the TreeSHAP explanations, the predictive utility is fully preserved ($0.8797 \pm 0.0138$ local CV, 0.8989 global test AUPRC).

Crucially, this privacy gain incurs only a modest reduction in explanation fidelity: DP-FedSHAP retains an 80.0\% Top-10 Feature Overlap and a strong positive Spearman Rank correlation of 0.793. Thus, 8 of the top 10 most influential fraud indicators remain intact, maintaining sufficient explanatory utility for human auditors while successfully obfuscating individual record signatures.

\subsection{Weight-Level DP: An Instructive Failure Mode}

To validate the design rationale of DP-FedSHAP, a Weight-Level DP baseline was evaluated under an identical privacy budget ($\epsilon = 1.2$, $C = 0.8$), but with noise applied to the trained model's leaf weights rather than to post-hoc explanation vectors. This configuration achieved an MIA accuracy of 49.92\%, statistically comparable to DP-FedSHAP's 49.75\%. However, its global test AUPRC collapsed to 0.5040, indistinguishable from random guessing on this balanced binary task. This is the central finding of the comparison: an identical privacy budget can neutralize the attacker equally well at either the weight level or the explanation level, but only one of these two placements preserves a usable model. Injecting DP noise into the structural components of a gradient-boosted tree ensemble, which relies on precise leaf thresholds to make useful splits, destroys the model's discriminative capacity entirely. A defense that eliminates membership leakage only by also eliminating predictive utility is not a viable deployment option for a fraud detection system. By contrast, DP-FedSHAP achieves comparable privacy protection while preserving the model's full predictive utility (AUPRC = 0.8989), because the perturbation is confined to the explanation layer and never touches the decision boundary itself. Weight-Level DP also degrades explanation fidelity more severely than DP-FedSHAP, retaining only a 70.0\% Top-10 Feature Overlap and a Spearman Rank of 0.7623, compared to DP-FedSHAP's 80.0\% and 0.793 respectively -- a further indication that noising the model's internal structure disrupts its behaviour less predictably than noising the explanation layer directly.

\begin{table}[htbp]
\caption{Privacy--Utility--Fidelity Trade-off Across Architectures}
\label{tab1}
\begin{center}
\begin{tabular}{|l|c|c|c|c|}
\hline
\textbf{Architecture} & \textbf{AUPRC} & \textbf{MIA Acc.} & \textbf{Top-10} & \textbf{Spearman} \\
\hline
Centralized Baseline        & $0.9012 \pm 0.0031$       & 61.67\% & 100.0\% & 1.000 \\
Standard Federated          & 0.8989\textsuperscript{*} & 51.58\% & 100.0\% & 0.980 \\
DP-FedSHAP ($\epsilon=1.2$) & 0.8989\textsuperscript{*} & 49.75\% &  80.0\% & 0.793 \\
Weight-Level DP ($\epsilon=1.2$) & 0.5040\textsuperscript{*} & 49.92\% & 70.0\% & 0.7623 \\
\hline
\end{tabular}
\end{center}
\footnotesize\textsuperscript{*}Single-split global ensemble estimate; see Limitations.
\end{table}

\subsection{Practical Implications}

These findings demonstrate that while federated model aggregation reduces direct membership signals, high-fidelity explanation outputs remain an open attack vector. Applying calibrated, post-hoc DP to the explanation layer resolves this vulnerability without requiring intrusive architectural changes or sacrificing model predictive performance. The Weight-Level DP comparison further shows that the choice of \emph{where} in the pipeline DP noise is injected is not incidental: applying an identical privacy budget to the model's internal structure, rather than to its explanations, can neutralize an attacker at the cost of the model's core predictive function, whereas explanation-level DP achieves comparable privacy protection with no such trade-off.

\subsection{Limitations}

This study has several limitations. The federated architecture was simulated across three client nodes under IID partition assumptions; heterogeneous (non-IID) client distributions warrant further study. While the centralized baseline and client-level training were validated using stratified 5-fold cross-validation, the aggregated federated ensemble and downstream MIA evaluation are reported on a single stratified test partition. The DP-FedSHAP and Weight-Level DP mechanisms compute a shared clipping threshold across member and non-member batches and use a fixed random seed for noise generation, resolving an earlier instability in which independently-thresholded batches produced MIA accuracy estimates ranging from approximately 20\% to 52\% across repeated runs. The reported values reflect a single seeded run; averaging MIA accuracy over multiple independently seeded noise draws remains recommended future work to further characterize the stability of the reported estimates. Furthermore, the threat model evaluates a non-adaptive adversary; testing against an attacker who retrains against noisy explanations remains an avenue for future work. Finally, the dataset was fully rebalanced to a 1:1 fraud ratio prior to training; the sensitivity of these results to the natural class imbalance of the underlying IEEE-CIS dataset (approximately 3.5\% fraud prevalence) was not directly tested and is identified as a direction for future work.

\section{Conclusion}

This dissertation demonstrates that high-fidelity TreeSHAP explanations leak sensitive membership information in financial fraud detection systems, and that standard federated averaging alone does not completely eliminate this risk. Four findings support this conclusion. First, the centralized baseline confirmed that raw SHAP vectors are directly exploitable, yielding an MIA accuracy of 61.67\% against a random-guess baseline of 50\%. Second, standard federated averaging alone reduced this leakage to 51.58\%, an improvement driven purely by the incidental averaging of client explanations rather than any formal privacy mechanism, and therefore an unreliable defense on its own. Third, the proposed DP-FedSHAP architecture closed this gap, reducing MIA accuracy to 49.75\% while preserving an AUPRC of 0.8989 and an 80.0\% Top-10 feature overlap relative to the undefended explanations. Fourth, a matched Weight-Level DP baseline, using an identical privacy budget but noising the trained model's leaf weights instead of its explanations, achieved a statistically comparable MIA accuracy of 49.92\%, but collapsed predictive utility to near-random performance (AUPRC = 0.5040), directly demonstrating that the placement of DP noise in the pipeline determines whether privacy protection preserves or destroys the model's usefulness.

The central contribution of this work is architectural: by applying $L_2$ norm sensitivity bounding and calibrated Laplacian noise directly to post-hoc explanation vectors, rather than to the trained model's internal structure, DP-FedSHAP avoids the utility collapse observed when the same privacy budget is applied at the weight level in tree-based ensembles. This makes formal privacy protection practically deployable for financial institutions that rely on XGBoost and TreeSHAP for regulatory transparency, without requiring a change to the underlying predictive model.

Future work will extend this evaluation along four directions. First, MIA accuracy could be averaged over multiple independently seeded noise draws to further characterize the stability of the reported privacy estimates. Second, adaptive clipping thresholds could be investigated to optimize explanation fidelity under streaming, non-IID client conditions, where fraud patterns and data volumes vary across institutions. Third, validating DP-FedSHAP against additional financial fraud benchmarks with differing feature distributions, and against the dataset's natural class imbalance rather than a fully rebalanced sample, would strengthen confidence in its generalizability. Fourth, evaluating the mechanism against adaptive adversaries who retrain their attack strategy in response to the defense would test the robustness of the empirical MIA results reported here under a stronger threat model.

\end{document}